\documentclass[conference]{IEEEtran}
\IEEEoverridecommandlockouts
\usepackage{cite}
\usepackage{amsmath,amssymb,amsfonts}
\usepackage{algorithm}  
\usepackage{algorithmicx}  
\usepackage{algpseudocode}
\usepackage{graphicx}
\usepackage{textcomp}
\usepackage{xcolor}
\usepackage{multirow}
\usepackage{bm}
\usepackage{hyperref}

\def\BibTeX{{\rm B\kern-.05em{\sc i\kern-.025em b}\kern-.08em
    T\kern-.1667em\lower.7ex\hbox{E}\kern-.125emX}}
\hypersetup{hidelinks}
\begin{document}

\title{Evolving Neural Networks through a Reverse Encoding Tree}

\author{
\small
\begin{tabular}[t]{cccc} 
\textbf{Haoling Zhang} & \textbf{Chao-Han Huck Yang} & \textbf{Hector Zenil}  &  \textbf{Narsis A. Kiani} \\
\textit{Institute of Biochemistry} & \textit{School of ECE} & \textit{Algorithmic Dynamics Lab \&} & \textit{Algorithmic Dynamics Lab} \\ 
\textit{BGI-Shenzhen} & \textit{Georgia Institute of Technology} & \textit{Oxford Immune Algorithmics} & \textit{Karolinska Institute} \\
Shenzhen, Guangdong, China & Atlanta, GA, USA & U.K. \& Sweden & Stockholm, Sweden \\
zhanghaoling@genomics.cn & huckiyang@gatech.edu &hector.zenil@cs.ox.ac.uk & narsis.kiani@ki.se 
\\ \\
\multicolumn{2}{c}{\textbf{Yue Shen}} & \multicolumn{2}{c}{\textbf{Jesper N. Tegner*}}\\
\multicolumn{2}{c}{\textit{Institute of Biochemistry}} & \multicolumn{2}{c}{\textit{Living Systems Lab, BESE, CEMSE}} \\
\multicolumn{2}{c}{\textit{BGI-Shenzhen}} & \multicolumn{2}{c}{\textit{King Abdullah University of Science and Technology}} \\
\multicolumn{2}{c}{Shenzhen, Guangdong, China} & \multicolumn{2}{c}{Thuwal 23955, Saudi Arabia} \\
\multicolumn{2}{c}{shenyue@genomics.cn} & \multicolumn{2}{c}{jesper.tegner@kaust.edu.sa}
\end{tabular}
}
\maketitle

\begin{abstract}
NeuroEvolution is one of the most competitive evolutionary learning strategies for designing novel neural networks for use in specific tasks, such as logic circuit design and digital gaming. 
However, the application of benchmark methods such as the NeuroEvolution of Augmenting Topologies (NEAT) remains a challenge, in terms of their computational cost and search time inefficiency. This paper advances a method which incorporates a type of topological edge coding, named Reverse Encoding Tree (RET), for evolving scalable neural networks efficiently. 
Using RET, two types of approaches -- NEAT with Binary search encoding (Bi-NEAT) and NEAT with Golden-Section search encoding (GS-NEAT) -- have been designed to solve problems in benchmark continuous learning environments such as logic gates, Cartpole, and Lunar Lander, and tested against classical NEAT and FS-NEAT as baselines. 
Additionally, we conduct a robustness test to evaluate the resilience of the proposed NEAT approaches. 
The results show that the two proposed approaches deliver an improved performance, characterized by (1) a higher accumulated reward within a finite number of time steps; (2) using fewer episodes to solve problems in targeted environments, and (3) maintaining adaptive robustness under noisy perturbations, which outperform the baselines in all tested cases. 
Our analysis also demonstrates that RET expends potential future research directions in dynamic environments.
 Code is available from \href{https://github.com/HaolingZHANG/ReverseEncodingTree}{https://github.com/HaolingZHANG/ReverseEncodingTree}.
\end{abstract}


\section{Introduction}
NeuroEvolution (NE) is a class of methods for evolving artificial neural networks through evolutionary strategies~\cite{stanley2019designing, zador2019a}.
The main advantage of NE is that it allows learning under conditions of sparse feedback. 
In addition, the population-based process makes for good parallelism~\cite{lehman2013neuroevolution}, without the computational requirement of back-propagation.
The evolutionary process of NE is modifying connection weights in the fixed network topology (individuals)~\cite{stanley2002evolving} by calculating their fitnesses and evaluating the relationship between individuals.

Recent studies~\cite{stanley2002evolving, whiteson2005automatic, lehman2018safe} show that the trade-off between protecting topological innovations and promoting evolutionary speed is also a challenge. The evolutionary process from the initial to the final individual is difficult to control accurately.
Genetic Algorithm (GA) using Speciation Strategies~\cite{knapp2019natural} allow a meaningful application of the crossover operation and protect topological innovations, avoiding premature disappearance.
The distribution estimation algorithms, such as Population-Based Incremental Learning~\cite{holker2010toward} (PBIL), represents a different way of describing the distribution information of candidate topologies of neural networks in the search space, i.e. by establishing a probability model.
The Covariance Matrix Adaptation Evolution Strategy~\cite{hansen2001completely} (CMA-ES) further explains the correlations between the parameters of a targeted fitness function, correlations which significantly influence the time taken to find a suitable control strategy~\cite{igel2003neuroevolution}. 
Safe Mutation~\cite{lehman2018safe} can scale the degree of mutation of each weight, and thereby expand the scope of domains amenable to NE.
In this study, a mapping relationship, based on constraining the topological scale, is set up between its features and fitness, in order to explore how the evolutionary strategy influences the population in a restricted search space.
The limitation of this scale serves to prevent unrestricted expansion of structures of neural network during evolutionary process.
On the restricted topological scale, all neural networks that can be generated by their features have achieved fitness through specific tasks.
The location of the specific neural network is the location of its features on the constrained topological scale.
In this situation, the location of the nearest two neural networks can be regarded as infinitesimal, and the function made up of all locations is continuous.
We define the location of an individual, which created by its feature matrix, as the input of the function, and its fitness as the output. 
Together, all locations form a complex and smooth fitness landscape~\cite{wang2018vine}.

In this fitness landscape, all evolutionary processes of the topology of the neural network can be regarded as processes of tree-based searching, like random forest~\cite{liaw2002classification}.
The initial population can be regarded as the root nodes, and the population of each generation can be regarded as the branch nodes of each layer.
Based on the current population or other population information (such as the probability matrix), more representative or better nodes will be identified in the next layer and used as individuals in the next generation.
Interestingly, certain classical search methods have attracted our attention.
Some global search methods, like Binary Search~\cite{mussmann2018generalized} and Golden-Section Search~\cite{chang2009n}, are not merely of use in finding extreme values in uni-modal functions, but have also shown promise when used in other fields~\cite{ koupaei2016new, guillot2019search}.
The search processes of the above global search method are similar to the reverse process of tree-based searching~\cite{henikoff1994position}.
The final (root) node is dependent on the elite leaf or branch nodes in each layer, as the topology of the final neural network is influenced by the features of the elite topology of each generation.

Based on the reverse process of tree-based searching (as the evolutionary strategy), we design two specific strategies in the fitness landscape, named NEAT with a reverse binary encoding tree (Bi-NEAT) and NEAT with a reverse golden-section~\cite{kiefer1953sequential} encoding tree (GS-NEAT).
In addition, the correlation coefficient~\cite{emerson2015causation} is used to analyze the degree of exploration of multiple sub-clusters~\cite{saxena2017review} in the fitness landscape formed by each generation of individuals.
It effectively prevents the population from falling into the optimal local solution due to over-rapid evolution.
The evolution speed of NEAT and FS-NEAT (as the baselines) and our proposed strategies are discussed in the logic operations and continuous control gaming benchmark in the OpenAI-gym ~\cite{brockman2016openai}.
These strategies have also passed different levels and types of noise tests to establish their robustness.
We reach the following conclusions: (1) Bi-NEAT, and GS-NEAT can improve the evolutionary efficiency of the population in NE; (2) Bi-NEAT and GS-NEAT show a high degree of robustness when subject to noise; (3) Bi-NEAT and GS-NEAT usually yield simpler topologies than the baselines.

\section{Related Work}
In this study, we introduce a search method into NeuroEvolution, and extract features in the neural network for the purpose of encoding the feature matrix.
Therefore we devote this section to brief descriptions of the following three topics: (1) Evolutionary Strategies in NeuroEvolution; (2) Search Methods; and (3) Network Coding Methods.

\subsection{Evolutionary Strategies in NeuroEvolution}
NeuroEvolution (NE) is a combination of Artificial Neural Network and Evolutionary Strategy.
Preventing crossovering the topologies efficiently, protecting new topological innovation, and keeping topological structure simple are three core problems faced in dealing with the Topology and WEight of Artificial Neural Network (TWEANN) system~\cite{reisinger2004evolving}.
In recent years, many effective ideas have been introduced into NE.
An important breakthrough came in the form of NEAT~\cite{stanley2002evolving, whiteson2005automatic}, which protects innovation by using a distance metric to separate networks in a population into species, while controlling crossover with innovation numbers.
However, the evolutionary efficiency of the population in each generation cannot be guaranteed.
In order to guarantee the evolutionary efficiency of NE, three research paths have been devised: (1) the replacement of the original speciation strategy with a new speciation strategy~\cite{knapp2019natural}; (2) the introduction of more effective evolutionary strategies~\cite{igel2003neuroevolution, holker2010toward, lehman2018safe}; (3) the use of novel topological structures~\cite{watts2019blocky}.

Certainly, modifying the structure and/or weight involves much more than the feature information of ANN itself.
The above improvements make it challenging to prevent the modification of all features.
Furthermore, the complexity of the topology required for obtaining the required ANN is unlimited, which means that the topological structure of ANN will not be necessarily simple.

\subsection{Search Methods}
In the field of Computer Science, search trees, such as the Binary Search tree~\cite{bentley1975multidimensional}, are based on the idea of divide and conquer method. 
They are often used to solve extrema in uni-modal arrays or find specific values in sorted arrays. 

Recently, some improved search trees have also been used to solve extrema in multi-modal or other optimization fields~\cite{southwell2012complex, koupaei2016new, tanyildizi2018novel, guillot2019search}.
These search trees, such as Binary Search~\cite{mussmann2018generalized}, Golden-Section Search~\cite{chang2009n}, and Golden-Section SINE Search ~\cite{tanyildizi2018novel} complete complex tasks by combining with population ~\cite{aurasopon2019improved} or other strategies~\cite{koupaei2016new}.
They make the whole population develop more accurately with geometric searching.
In the field of multi-modal searching~\cite{das2011real}, they increase global optimization ability by estimating and abandoning small peaks.
Therefore using tree-based search has the potential to improve evolutionary efficiency.
In addition, tree-based searches have a strong resistance to environmental noise~\cite{wang2019fibonacci}, where position of optimum point would be generated by a sampling-based distribution to enhance interference on noisy observation.

Given the crossover operation of topologies, some search methods have spurred an interest in enhancing the precision of such crossover operations, thus opening up an interesting avenue for the introduction of search trees into NE.

\subsection{Network Coding Methods}
At the stage of direct coding, the encoding rule of ANN is to convert it into a genotype~\cite{stanley2002evolving}.
In order to generate large-scale, functional, and complex ANN, some indirect coding~\cite{auerbach2011evolving, huizinga2016does} techniques have been proposed.
However, they are not efficient enough for the evolution of local networks, because decreasing the granularity of coordinates leads to a decrease in resolution~\cite{stanley2007compositional}.
The above encoding is a kind of cellular encoding~\cite{gruau1992genetic}, which uses chromosomes or genotypes consisting of trees of node operators to evolve a graph.

Edge encoding~\cite{luke1996evolving}, which is different from cellular encoding, grows a graph by modifying its edges, and thus has a different encoding bias than the genetic search process.
When naturally evolving network topologies, edge encoding is often better than cellular encoding~\cite{luke1996evolving}.
Edge encoding can use adjacency matrices as representational tools \cite{brisaboa2009k}.
An adjacency matrix represents a graph with a type of square matrix where each element represents an edge. 
The corresponding nodes connected by weight are indicated by the row and column of the edge in the matrix.

\section{NeuroEvolution of Reverse Encoding Tree}
We propose an advanced search method, named Reverse Encoding Tree (RET), to leverage the existing speciation strategy~\cite{stanley2002evolving} in NEAT.
The edge encoding~\cite{luke1996evolving} with the adjacency matrix is the representation of RET for network coding. 
RET uses unsupervised clusters~\cite{jin2004reducing} to dynamically describe speciation and speciation relationships.
To reduce the complexity of the terminated network~\cite{reisinger2004evolving}, RET limits the maximum number of nodes in all generated neural networks.

\begin{figure}[htbp]
\centerline{\includegraphics[width=0.95\columnwidth]{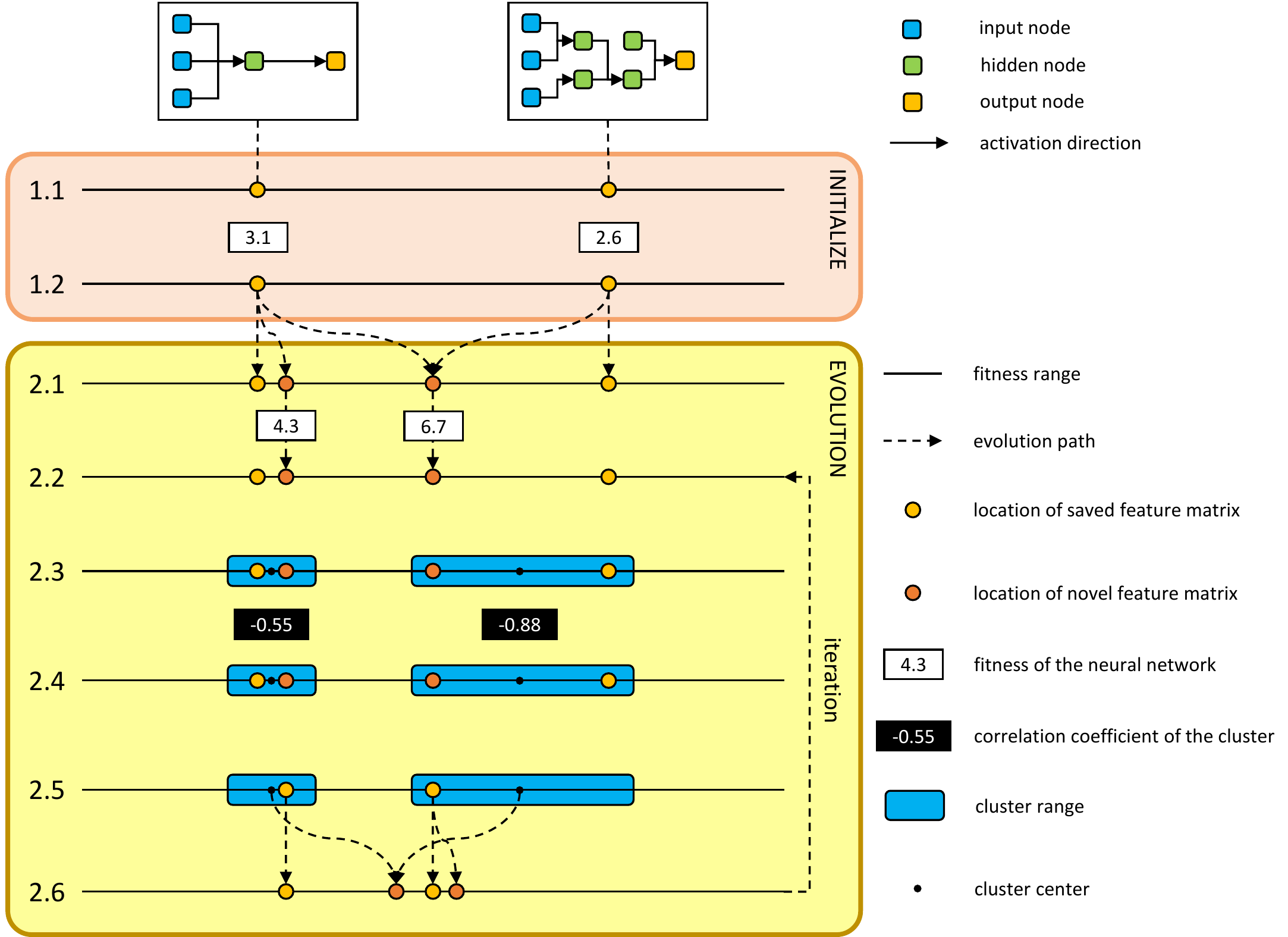}}
\caption{Flowchart of Bi-NEAT, a specific strategy in RET.
The detailed description two internal processes are as follow: (1.1) create the first generation globally in the $fitness$ landscape; (1.2 and 2.2) calculate fitness of neural network; (2.1) build second generation by RET and calculate the fitnesses; (2.3) divide the current generation to the (population-size)-clusters; (2.4) calculate the correlation coefficient of each cluster; (2.5) save the best genome each cluster as the next generation; (2.6) create the novel genomes based on RET as the next generation.}
\label{fig:flowchart}
\end{figure}

An illustration of this strategy (using binary search, namely Bi-NEAT), is provided in Fig.~\ref{fig:flowchart}.
Different from the speciation strategies in NEAT, RET crosses
topologies by search method and evaluates the relationships within and between species by best fitness and correlation coefficient in each cluster, which estimates the small peaks in the $fitness$ landscape.
Through abandoning these small peaks, RET speeds up the evolutionary process of NE.

\subsection{Network Encoding}
The evolution of the neural network can be achieved by changing network structure, connection weight, and node bias.
Changing the topology of neural networks is a coarse-grained evolutionary behavior~\cite{maniezzo1994genetic}.
Therefore, to search for the solution space more smoothly, we first limit the maximum number of nodes ($m$) in the neural network.
The explorable range of the population is therefore fixed and limited to avoid unrestricted expansion of the topology of the neural network during the evolutionary process.
The limitation of nodes generated would give the weight and bias information in the specific network a greater chance of being optimized.

We first introduce a $fitness$ landscape ($\Theta$) as a combination of generated neural networks with a fitness evaluation to perform a task in a targeted environment 
(e.g., XOR Gate or Cartpole~\cite{brockman2016openai}).
$\Theta$ includes all networks in the solution space.

We define a $individual$ seeding ($\bm{I}$) from the initial population in the range of $\Theta$ with a specified number ($p$), as $\bm{P} = \left \langle \bm{I}^{(0)}, \bm{I}^{(1)}, \ldots, \bm{I}^{(p)} \right \rangle$.
There is an initial distance ($d_i$) between each of the two genotypes, to ensure that the initial population can attain as much diversity as possible in $\Theta$.
In addition, the related hyper-parameter $d_m$ describes the minimum distance between two individuals. From previous studies~\cite{hoos2004stochastic}, it is known that $d_m$ reduces the efforts of the population to over-explore the local landscape.
The dynamics of individual would increase when the distance between a novel individual and other, existing individuals is less than $d_m$. The distance check equation is shown as:

\begin{equation} 
\label{eq:check}
{\rm check}(d, \bm{I}, \bm{P})
\end{equation}

\begin{figure}[htb]
\centerline{\includegraphics[width=0.4\columnwidth]{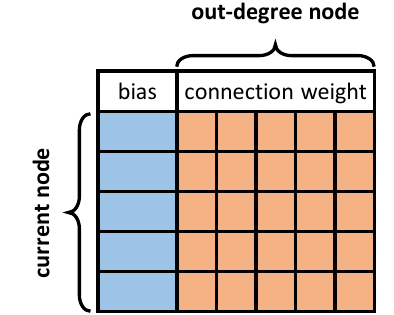}}
\caption{Feature matrix of the individual.}
\label{fig:feature_matrix}
\end{figure}

The distance between two individuals is encoded as the Euclidean distance~\cite{anton2013elementary} of the corresponding feature matrix:

\begin{equation} 
\label{eq:distance}
d(\bm{f}_i, \bm{f}_j)=\sqrt{\sum_{v \in {\rm det}(\bm{f}_i - \bm{f}_j)} v^2}
\end{equation}
where $\bm{f}$ is the feature matrix, in the range of $\Theta$.
In the feature matrix, the first column is the bias of each node, and the other columns are the connection weights between nodes in the neural network generated by the individual.
An illustration of the feature matrix is provided in Fig.~\ref{fig:feature_matrix}.
The feature information includes input, output, and hidden nodes.
Therefore, the size of the feature matrix is $m \times (m + 1)$.
Because the feature matrix includes all features of the individual, any individual can be created from its feature matrix by $\bm{I} = {\rm create}(\bm{f})$. 

\subsection{Evolutionary Process}
The population in the current generation is composed of the individuals saved (elite) from the population in the previous generation and the novel individuals generated by RET based on the landscape of the population in the previous generation. 

\begin{figure}[htbp]
\centerline{\includegraphics[width=0.8\columnwidth]{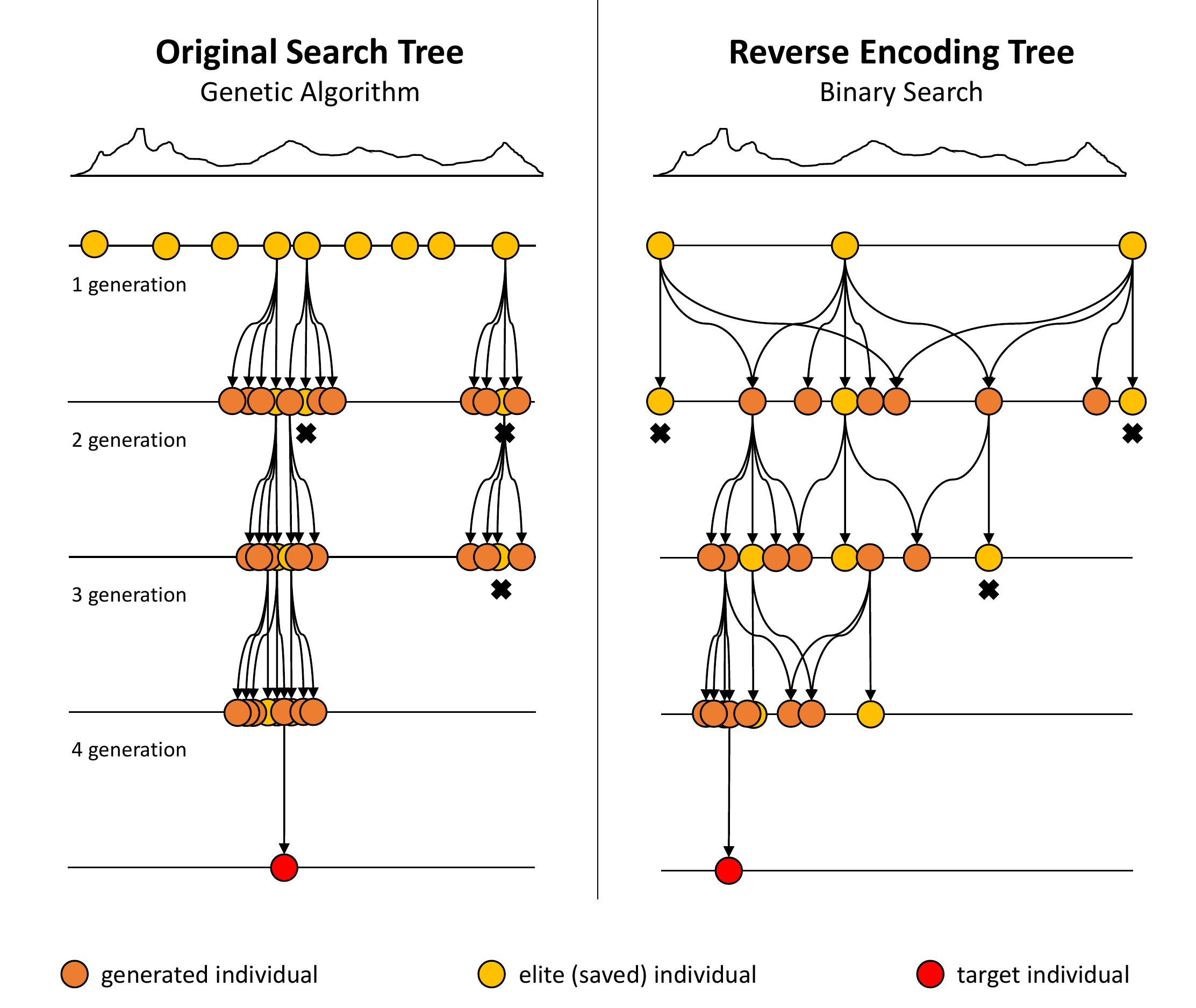}}
\caption{Illustration of two types of proposed tree-based network encoding.
In addition to original generated way of novel individuals, RET provides discovered opportunities for potential elite individuals on the global range, using edge encoding from every two elite individuals.}
\label{fig:ret}
\end{figure}

RET is different from original evolutionary strategies, as is shown in Fig.~\ref{fig:ret}. 

The novel adjacent individuals created by the elite individuals in the original tree search.
In RET, except for the above search, novel individuals also inserted as the root nodes of a tree by edge encoding from elite individuals (as the leaf nodes) for preserving all features between every two elite individuals and evaluating the global fitness landscape.

The search process of RET is divided into two parts: (1) the creation of a nearby individual from the specified parent individual by the original frame of NEAT:
\begin{equation} 
\label{eq:near}
\bm{I}^{(i)}_{\rm N} = {\rm neat}(\bm{I}^{(i)})
\end{equation} 
(2) the creation of a global individual from the two specified parent individuals or feature matrices:
\begin{equation} 
\label{eq:global}
\bm{I}^{(i,j)}_{\rm G} = 
\left\{
\begin{aligned}
\bm{I}^{(i,j)}_{\rm Bi}\\
\bm{I}^{(i,j)}_{\rm GS}
\end{aligned}
\right.
\end{equation} 

And this work includes binary search (Eq.~\ref{eq:bi}) and golden section search (Eq.~\ref{eq:gs}).

\begin{equation} 
\label{eq:bi}
\bm{I}^{(i,j)}_{\rm Bi} = {\rm create}(\frac{\bm{f}^{(i)} + \bm{f}^{(j)}}{2})
\end{equation} 

\begin{equation} 
\label{eq:gs}
\bm{I}^{(i,j)}_{\rm GS} = {\rm create}(\frac{(\sqrt{5} - 1) \times \bm{f}^{(i)} + (3 - \sqrt{5}) \times \bm{f}^{(j)}}{2})
\end{equation} 

\subsection{Analysis of Evolvability}
We further propose an efficient, unsupervised learning method for analyzing the network seeds generated.
The motivation for clustering the population~\cite{jin2004reducing} based on the similarity of individuals is to explore the evolvability of each type of individual set after protecting topology innovations.
The current population is divided into $p$ clusters for understanding the local situation of the landscape generated by the current population.
Many clustering methods can be used in this strategy. We compared K-means++~\cite{bachem2016approximate}, Spectral Clustering~\cite{yan2009fast}, and Birch Clustering~\cite{zhang1996birch}, and selected the most advanced, K-means++, thus:

\begin{equation} 
\label{eq:kmeans}
\begin{aligned}  
\bm{S}
&= {\rm cluster}(\bm{P}) \\
&= \arg \min \sum_{i=1}^p \sum_{\bm{I} \in \bm{S}^{(i)}} d(\bm{I.f}, \bm{c}^{(i)})
\end{aligned}
\end{equation}
where $\bm{S}$ is the set of $p$ clusters, $\bm{S}^{(i)}$ is the $i^{th}$ cluster, and $\bm{c}^{(i)}$ is the center of the $i^{th}$ cluster. The optimal individual $\bm{I}^{(i.m)}$ in the $i^{th}$ cluster can be obtained by comparing the fitness of each individual:

\begin{equation} 
\label{eq:gmax}
\bm{I}^{(i.m)} = \arg \max_{\bm{I} \in \bm{S}^{(i)}} \bm{I}.r
\end{equation}
where $r$ is the fitness of the individual.
The set of saved individuals collects the optimal individual in every cluster:

\begin{equation} 
\label{eq:saved}
\bm{P}_{\rm saved} = \langle \bm{I}^{(i.m)}, i \in \mathbb{N}^+ \cap i \leq p  \rangle
\end{equation} 

The correlation coefficient ($\rho$) of distance from the optimal position of the individual and fitness for all the individuals in each cluster is calculated, to describe the situation of each cluster:

\begin{equation} 
\label{eq:corr}
\begin{aligned}  
\rho^{(i)} 
&= {\rm corr}(\bm{I}^{(i.m)}, \bm{S}^{(i)}) \\
&= {\rm corr}(d(\bm{I}^{(i.m)}.\bm{f}, \bm{I.f}) \Longleftrightarrow \bm{I}.r, \bm{I} \in \bm{S}^{(i)})
\end{aligned} 
\end{equation} 

\begin{figure}[htb]
\centerline{\includegraphics[width=0.75\columnwidth]{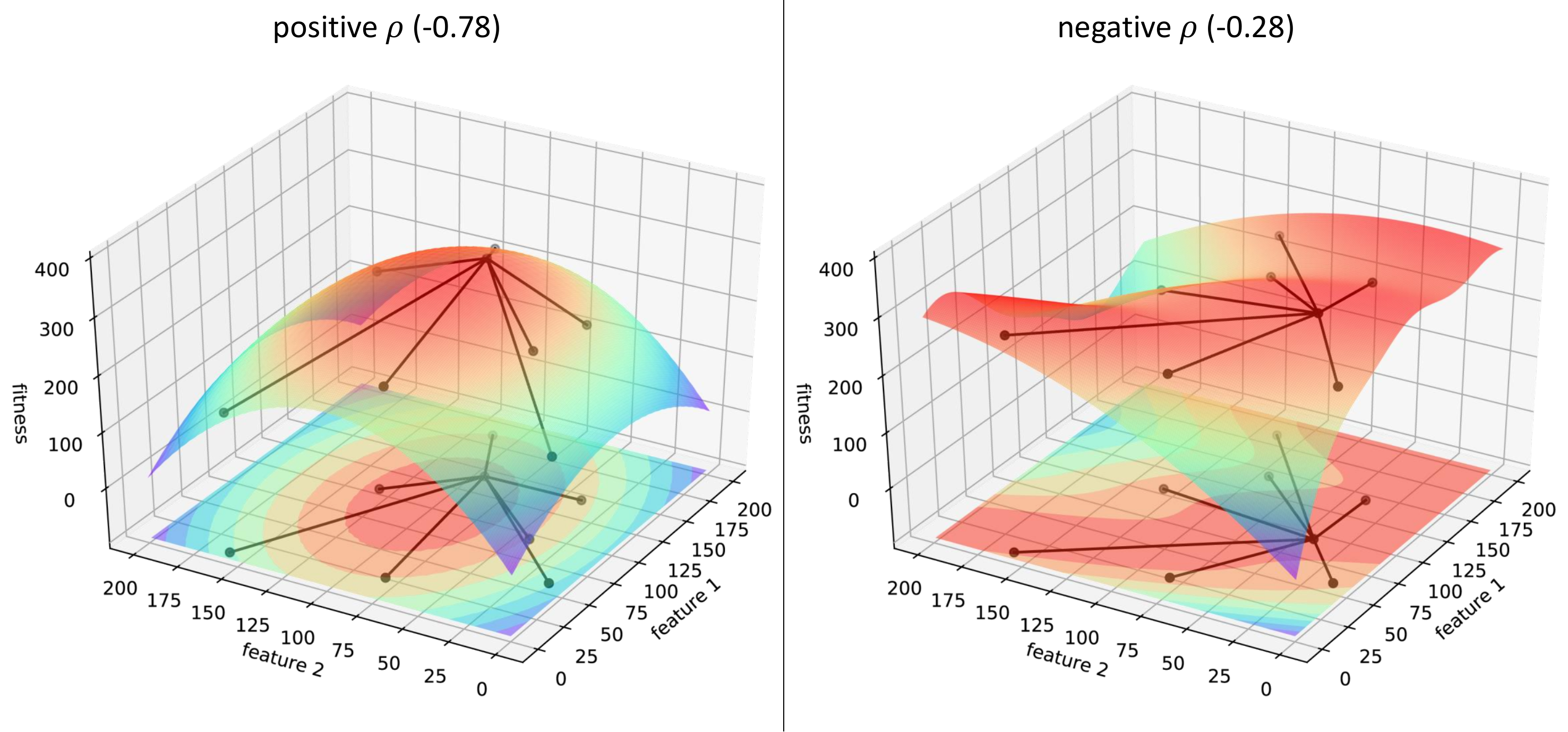}}
\caption{Two types of $\rho$ in a cluster}
\label{fig:corr}
\end{figure}

For the local $fitness$ landscape of a single maximum value, the distance and fitness show a negative correlation (positive $\rho$), $\rho$ will reach $-1$.
If the landscape is complex (negative $\rho$), the relationship between distance and fitness is not significant. 
Two types of $\rho$ are shown in Fig.~\ref{fig:corr}.

RET's operation occurs between each of the two clusters:

\begin{equation} 
\label{eq:novel}
\begin{aligned}  
\bm{P}_{\rm novel} 
&= \langle \sum_{i=1}^{p-1} \sum_{j=i+1}^{p} {\rm ret}_{ij} \rangle \\
&= \langle \sum_{i=1}^{p-1} \sum_{j=i+1}^{p} {\rm ret}( [\bm{S}^{(i)}, \rho^{(i)}], [\bm{S}^{(j)}, \rho^{(j)}]) \rangle
\end{aligned}  
\end{equation} 

The operation selection is dependent on the optimal individuals and the correlation coefficients of the two specified clusters.
Therefore, the number of novel individuals is less than or equal to $p^2 - p$.
We assume that if $\rho^{(i)} \leq -0.5$, $i^{th}$ cluster has been explored fully, or its local $fitness$ landscape is simple.
When $\bm{I}^{(i.m)} > \bm{I}^{(j.m)}$, the operation selection in each comparison is: 

\begin{equation} 
\label{eq:novel_ret}
{\rm ret}_{ij} =
\left\{
\begin{array}{cc}
\bm{I}^{(i.m)}_{\rm N}, \bm{c}^{(i, j)}_{\rm G} & {\rho^{(i)} \leq -\frac{1}{2} \cap \rho^{(j)} \leq -\frac{1}{2}}\\
\bm{I}^{(i.m)}_{\rm N}, \bm{I}^{(i.m)}_{\rm N} & {\rho^{(i)} > -\frac{1}{2} \cap \rho^{(j)} \leq -\frac{1}{2}} \\
\bm{I}^{(i.m)}_{\rm N}, \bm{I}^{(j.m)}_{\rm N} & {\rho^{(i)} \leq -\frac{1}{2} \cap \rho^{(j)} > -\frac{1}{2}} \\
\bm{I}^{(i.m)}_{\rm N}, \bm{I}^{(j.m)}_{\rm N} & {\rho^{(i)} > -\frac{1}{2} \cap \rho^{(j)} > -\frac{1}{2}}
\end{array}
\right.
\end{equation} 
where $\bm{c}^{(i, j)}_{\rm G}$ is the novel individual created by two centers of the specified cluster.

In summary, our proposed evolutionary strategy uses RET based on the local $fitness$ landscape to evolve the feature matrix of individuals in the population. 
The pseudo-code of this evolutionary process is shown in Alg.~\ref{code:process}.

\begin{algorithm}  
\caption{Evolution process of NEAT with RET}
\label{code:process}
\begin{algorithmic}[1]  
\Require $d_i$, $d_m$, $p$, $m$
\Ensure $\bm{I}$ 
\State $\bm{P} \gets \emptyset$
\While{${\rm len}(\bm{P}) < p$}
\State $\bm{I} \gets {\rm create}(\bm{f})$ where $\bm{f} \in \Theta$
\If{${\rm check}(d_i, \bm{I}, \bm{P})$}
\State $\bm{P} \gets \bm{P}  + \bm{I}$
\EndIf
\EndWhile
\While{True}
\State calculate $r$ in each $\bm{I}$ where $\bm{I} \in \bm{P}$
\If {one of $\bm{I}.r$ meet fitness threshold}
\State \Return{$\bm{I}$ where $\bm{I}.r$ meet fitness threshold}  
\EndIf
\State $\bm{S} \gets {\rm cluster}(\bm{P})$, $\bm{P}_{\rm saved} \gets \emptyset$, $\bm{P}_{\rm novel} \gets \emptyset$
\For{$i = 1 \to p$}  
\State $\rho^{(i)} \gets {\rm corr}(\bm{I}^{(i.m)}, \bm{S}^{(i)})$ 
\EndFor 
\For{$i = 1 \to p$}  
\If{${\rm check}(d_m, \bm{I}^{(i.m)}, \bm{P}_{\rm saved})$}
\State $\bm{P}_{\rm saved} \gets \bm{P}_{\rm saved}  + \bm{I}^{(i.m)}$
\EndIf
\EndFor
\For{$i = 1 \to (p - 1)$}
\For{$j = i + 1 \to p$}
\State $\bm{I}^{(1)}, \bm{I}^{(2)} \gets {\rm ret}( [\bm{S}^{(i)}, \rho^{(i)}], [\bm{S}^{(j)}, \rho^{(j)}])$
\If{${\rm check}(d_m, \bm{I}^{(1)}, \bm{P}_{\rm saved} + \bm{P}_{\rm novel})$}
\State $\bm{P}_{\rm novel} \gets \bm{P}_{\rm novel} + \bm{I}^{(1)}$
\EndIf
\If{${\rm check}(d_m, \bm{I}^{(2)}, \bm{P}_{\rm saved} + \bm{P}_{\rm novel})$}
\State $\bm{P}_{\rm novel} \gets \bm{P}_{\rm novel} + \bm{I}^{(2)}$
\EndIf
\EndFor
\EndFor
\State $\bm{P} \gets \bm{P}_{\rm saved} + \bm{P}_{\rm novel}$
\EndWhile
\end{algorithmic}  
\end{algorithm} 

\section{Experiments}
In order to verify whether NE based on tree search can improve evolutionary efficiency and fight against environmental noise effectively, we designed a two-part experiment: 
(1) We explore the effect of our proposed strategies and the baseline strategies in classical tasks, such as the logic gate; 
(2) We explore the effect of our proposed strategies and the baseline strategies in one of the classical tasks (Cartpole-v0) under different noise conditions as a robustness test for continuous learning~\cite{yang2020enhanced}. 

\subsection{Logic Gate Representative}
The two-input symbolic logic gate, XOR, is one of the benchmark environments in the NEAT setting. The task is to evolve a network that distinguishes a correct Boolean output from $\left\{True (1), False (0)\right\}$. 
The initial reward is $4.0$, and the reward will decrease by the Euclidean distance between ideal outputs and actual outputs.
We select a higher targeted reward of $3.999$ to tackle this environment.
In addition, we add three kinds of additional logic gate, IMPLY, NAND, and NOR, to explore algorithm performance with different task complexities.
The complete hyper-parameter setting in the logical experiments is as shown in Tab.~\ref{tab:logical_h}.
To enhance the reproducibility of our work, select the XOR environment from the most popular \textbf{neat-python}~\footnote{\url{https://neat-python.readthedocs.io/en/latest/xor_example.html}} package and open-source our implementation in the supplementary material.

\begin{table}[htbp]
\centering
\caption{Hyper-parameters in the logical experiments.}  
\label{tab:logical_h}
\begin{tabular}{|c|c|}
\hline
\textbf{hyper-parameter} & \textbf{value}\\
\hline
iteration & 1000 \\
\hline
fitness threshold & 3.999 \\
\hline
evolution size & 132 \\
\hline
activation & sigmoid \\
\hline
\end{tabular}
\begin{flushleft} 
{Iteration represents the repeated number on one experiment, the metrics is coherent with previous works~\cite{watts2019blocky, knapp2019natural}.
``evolution size'' describes the number of individuals need to be evolved structure and calculated fitness in each generation, which is different from population size~\cite{stanley2002evolving}.
}
\end{flushleft}
\end{table}

\subsection{Continuous Control Environment}
\begin{figure}[htb]
\centerline{\includegraphics[width=0.9\columnwidth]{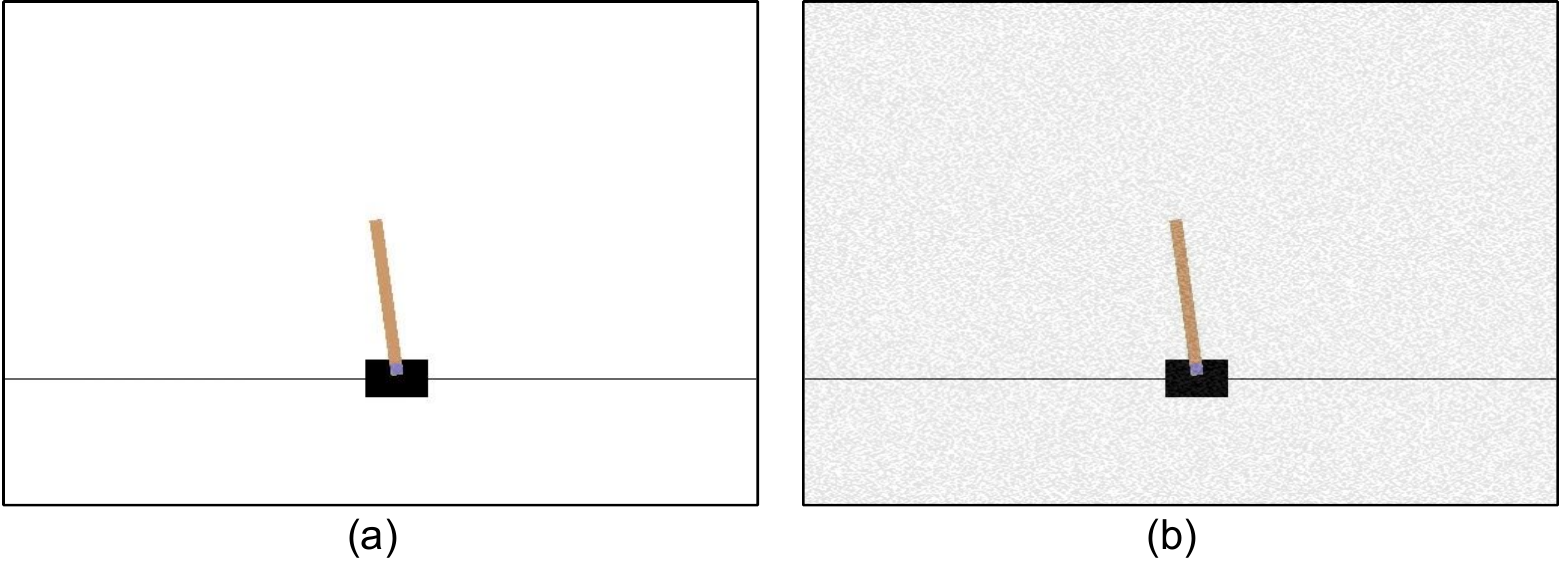}}
\caption{Illustration of a continuous control environment utilized as our task: (1) Cartpole-v1~\cite{brockman2016openai} and (2) Cartpole subject to a background perturbation of Gaussian noise.}
\label{fig:cartpole}
\end{figure}

Our testing platforms were based on OpenAI Gym~\cite{brockman2016openai}, well adapted for building a baseline for continuous control.
\textbf{Cartpole:}  As a classical continuous control environment~\cite{barto1983neuronlike}, the Cartpole-v0~\cite{brockman2016openai} environment is controlled by bringing to bear a force of $+1$ or $-1$ to the cart. A pendulum starts upright, and the goal is to prevent it from toppling over. An accumulated reward of $+1$ would be given before a terminated environment (e.g., falling $15$ degrees from vertical, or a cart shifting more than $2.4$ units from the center).
As experimental settings, we select $1000$ iteration, and use relu activation for neural network output to select an adaptive action in Tab.~\ref{tab:continuous_h}. To solve the problem, we conduct and fine-tune both NEAT and FS-NEAT as baseline results for accessing targeted accumulated rewards of $195.0$ in $200$ episode steps ~\cite{brockman2016openai}.

Here, we have improved the requirements of the fitness threshold ($499.5$ rewards in $500$ episode steps) and normalized the fitness threshold as $\frac{\rm rewards}{\rm episode\ steps}$. See Tab.~\ref{tab:continuous_h}.

\begin{table}[htbp]
\centering
\caption{Hyper-parameters in the Cartpole v0.}  
\label{tab:continuous_h}
\begin{tabular}{|c|c|}
\hline
\textbf{hyper-parameter} & \textbf{value}\\
\hline
iteration  & 1000 \\
\hline
fitness threshold & 0.999 \\
\hline
evolution size & 6 \\
\hline
activation & relu \\
\hline
episode steps & 500 \\
\hline
episode generation & 20 \\
\hline
\end{tabular}
\end{table}

\subsection{Gaming Environment}

\begin{figure}[htb]
\centerline{\includegraphics[width=0.9\columnwidth]{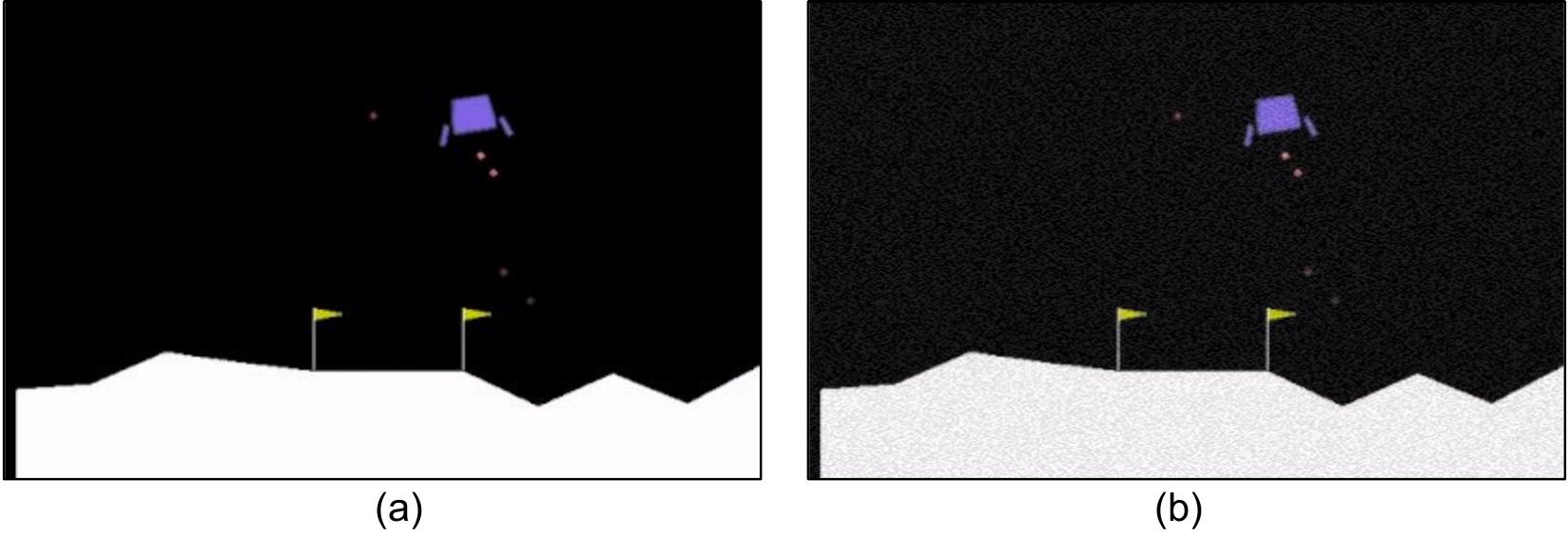}}
\caption{Illustration of a 2D gaming environment utilized as our task: (1) Lunar Lander-v2 from the OpenAI Gym~\cite{brockman2016openai} and (2) Lunar Lander-v2 subject to a background perturbation of Gaussian noise.}
\label{fig:lunar}
\end{figure}

\textbf{Lunar Lander:} We utilize a box-2d gaming environment, lunar lander-v2 as shown in Fig.~\ref{fig:lunar}, from OpenAI Gym~\cite{brockman2016openai}. 
The objective of the game is to navigate the lunar lander spaceship to a targeted landing site on the ground without collision, using two lateral thrusters and a rocket engine. 
Each episode lasts at most $1000$ steps and runs at $50$ frames per second. 
An episode ends when the lander flies out of borders, remains stationary on the ground, or when time is expired. 
A collection of six discrete \textbf{actions} that correspond to the $\left \{ left, right \right \}$ off steering commands and $\left \{ on, off \right \}$  main engine settings. 
The state, $s \in R^{8}$, is an eight-dimensional vector that continuously records and encodes the lander’s position, velocity, angle, angular velocity, and indicators for the contact between the legs of the vehicle and the ground. 
For the experiment, we run $1000$ iterations for the Cartpole-v0 setting with details in Tab.~\ref{tab:lunar_lander_h}.

\begin{table}[htbp]
\centering
\caption{Hyper-parameters in the LunarLander v2.}  
\label{tab:lunar_lander_h}
\begin{tabular}{|c|c|}
\hline
\textbf{hyper-parameter} & \textbf{value}\\
\hline
iteration & 1000 \\
\hline
fitness threshold & -0.2 \\
\hline
evolution size & 20 \\
\hline
activation & relu \\
\hline
episode steps & 100 \\
\hline
episode generation & 2 \\
\hline
\end{tabular}
\end{table}

\subsection{Robustness}
\label{sec4:robust}
One of the remain challenges for continuous learning is noisy observation ~\cite{osband2019behaviour} in the real-world. We further evaluate the Cartpole-v0~\cite{brockman2016openai} with a shared noisy benchmark from the bsuite~\cite{osband2019behaviour}. 
The hyper-parameter setting is shown in Tab.~\ref{tab:noise_h}. 

\textbf{Gaussian Noise}
Gaussian noise or white noise is a common interference in sensory data. The interfered observation becomes $\mathcal S_t = s_t + n_t$ with a Gaussian noise $n_t$. We set up the Gaussian noise by computing the variance of all recorded states with a mean of zero. 

\textbf{Reverse Noise}
Reverse noise maps the original observation data reversely. 
Reverse noise is a noise evaluation for sensitivity tests with a higher L2-norm similarity but should affect the learning behavior on the physical observation.
Reverse observation has been used in the continuous learning framework for communication system~\cite{liu2018anti} to test its robustness against jamming attacks.
Since 100\% of the noise environment is consistent with a noise-free environment, we dilute the noise level to the original 50\% (as dilution coefficient in Reverse).

\begin{table}[htbp]
\centering
\caption{Hyper-parameters in the noise experiments.}  
\label{tab:noise_h}
\begin{tabular}{|c|c|}
\hline
hyper-parameter & value\\
\hline
benchmark task & CartPole v0 \\
\hline
iteration & 1000 \\
\hline
evolution size & 6 \\
\hline
activation & relu \\
\hline
episode steps & 300 \\
\hline
episode generation & 2 \\
\hline
normal maximum & 0.10 \\
\hline
normal minimum & 0.05 \\
\hline
dilution coefficient in Reverse & 50\% \\
\hline
peak in Gaussian & 0.20 \\
\hline
\end{tabular}
\end{table}

\subsection{Baselines}
Here we take NEAT and FS-NEAT as baselines.
The weight of connection and bias of node are the default settings in the example of \textbf{neat-python}.

\section{Results}
\begin{figure}[htbp]
\centerline{\includegraphics[width=0.95\columnwidth]{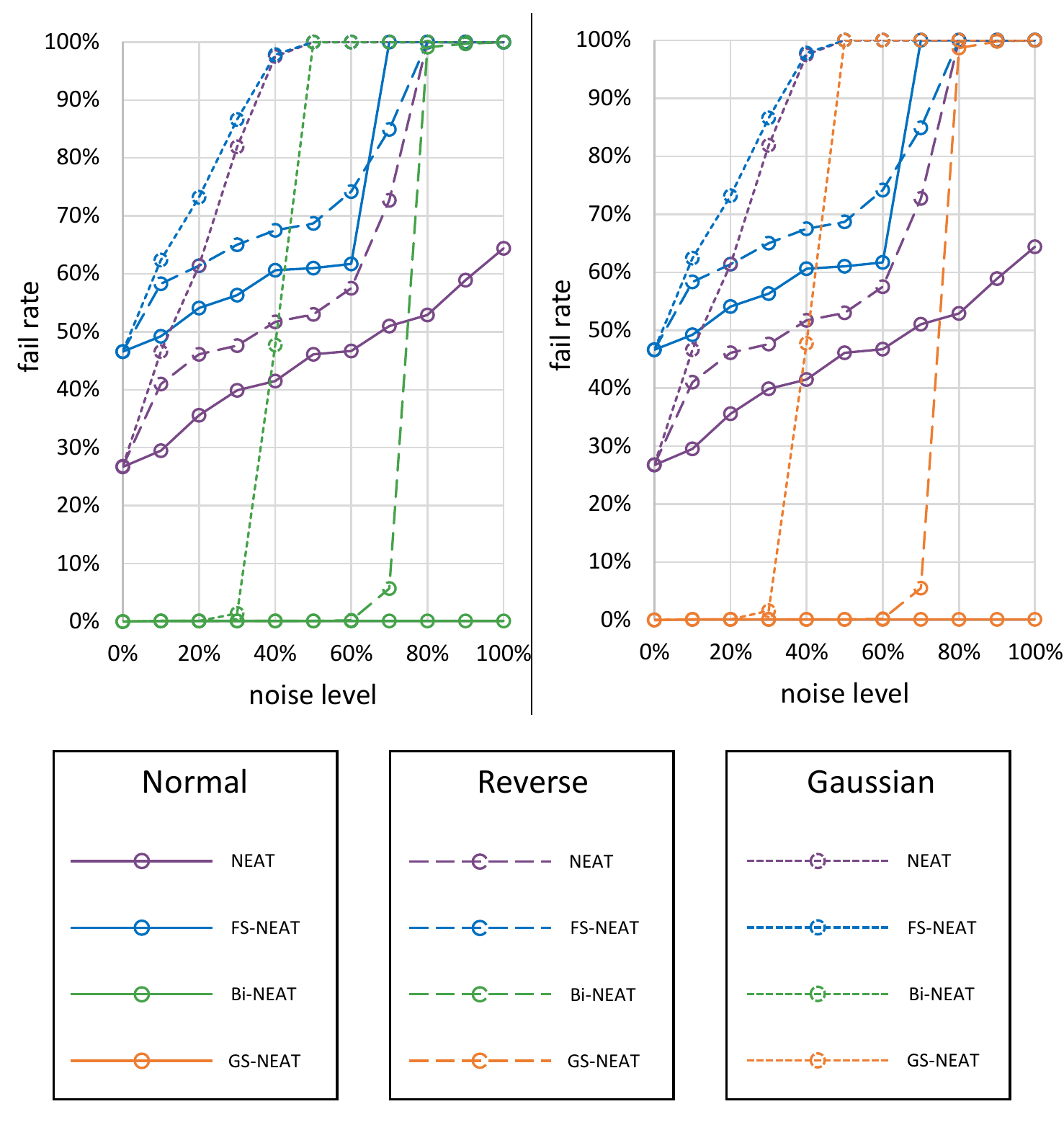}}
\caption{Robust evaluation in CartPole-v0 thorough noisy observations included: reverse and Gaussian perturbations in Sec.~\ref{sec4:robust}.}
\label{fig:attack}
\end{figure}

After running 1000 iterations for each method in the logical experiments, continuous control and game experiments, and noise attack experiments, we obtained the results shown in Tab.~\ref{tab:logical_r}, Tab.~\ref{tab:game_r}, and Fig.~\ref{fig:attack}.
The evolutionary process across all the methods has the same fitness number in each generation.
Therefore the comparison of average end generation is the same as the comparison of calculation times for the neural network in the evolutionary process.

After restraining the influence of hyper-parameters, the tasks from Tab.~\ref{tab:logical_r} describe the influence of task complexity on evolutionary strategies.
The results show that with the increase in task difficulty, our algorithm can make the population evolve faster.
In the IMPLY task, the difference between the average end generation is 1 to 2 generations.
When the average of end generations in XOR tasks is counted, the gap between our proposed strategies and the baselines widens to nearly 20 generations.
Additionally, the average node number in the final neural network and the task complexity seem to have a potentially positive correlation.

\begin{table}[htbp]
\centering
\caption{Result statistics in the experiments of logic gates.}  
\label{tab:logical_r}
\begin{tabular}{|c|c|c|c|c|}
\hline
task & method & fall rate & Avg.gen & StDev.gen \\
\hline
\multirow{4}{*}{IMPLY} 
& NEAT & 0.1\% & 7.03 & 1.96 \\
\cline{2-5}
& FS-NEAT & 0.0\% & 6.35 & 2.21 \\
\cline{2-5}
& Bi-NEAT & 0.0\% & 5.00 & 2.50\\
\cline{2-5}
& GS-NEAT & 0.0\% & 5.82 & 2.88\\
\hline
\multirow{4}{*}{NAND} 
& NEAT & 0.1\% & 13.02 & 3.87 \\
\cline{2-5}
& FS-NEAT & 0.0\% & 12.50 & 4.34 \\
\cline{2-5}
& Bi-NEAT & 0.0\% & 10.26 & 5.26\\
\cline{2-5}
& GS-NEAT & 0.0\% & 11.74 & 5.82\\
\hline
\multirow{4}{*}{NOR} 
& NEAT & 0.1\% & 13.13 & 4.18 \\
\cline{2-5}
& FS-NEAT & 0.0\% & 12.83 & 4.58 \\
\cline{2-5}
& Bi-NEAT & 0.0\% & 10.60 & 5.64\\
\cline{2-5}
& GS-NEAT & 0.0\% & 11.86 & 6.29\\
\hline
\multirow{4}{*}{XOR} 
& NEAT & 0.1\% & 103.42 & 56.02 \\
\cline{2-5}
& FS-NEAT & 0.1\% & 101.19 & 50.72 \\
\cline{2-5}
& Bi-NEAT & 0.0\% & 84.15 & 30.58 \\
\cline{2-5}
& GS-NEAT & 0.0\% & 88.11 & 36.13 \\
\hline
\end{tabular}
\end{table}

The tasks in the continuous control and game environments Bi-NEAT and GS-NEAT still show amazing potential. See Tab.~\ref{tab:game_r}.
Unlike in the case of the logical experiments, the results show that the two proposed strategies are superior both in terms of evolutionary speed and stability.
The enhanced evolutionary speed is reflected in the fact that the baselines require two to three times the average end generation as our strategies for the tested tasks.
In addition, the smaller standard variance of end generation shows the evolutionary stability of our strategies.

\begin{table}[htbp]
\centering
\caption{Result statistics in the complex experiments.}  
\label{tab:game_r}
\begin{tabular}{|c|c|c|c|c|}
\hline
task & method & fall rate & Avg.gen & StDev.gen \\
\hline
\multirow{4}{*}{CartPole v0} 
& NEAT & 26.5\% & 147.33 & 99.16 \\
\cline{2-5}
& FS-NEAT & 4.8\% & 72.86 & 85.08 \\
\cline{2-5}
& Bi-NEAT & 0.0\% & 29.35 & 18.86 \\
\cline{2-5}
& GS-NEAT & 0.0\% & 31.95 & 22.56 \\
\hline
\multirow{4}{*}{LunarLander v2} 
& NEAT & 4.9\% & 144.21 & 111.87 \\
\cline{2-5}
& FS-NEAT & 3.3\% & 152.91  & 108.61 \\
\cline{2-5}
& Bi-NEAT & 0.0\% & 48.66 & 44.57 \\
\cline{2-5}
& GS-NEAT & 0.0\% & 44.57 & 50.29 \\
\hline
\end{tabular}
\end{table}

As shown in Fig.~\ref{fig:attack}, the evolutionary strategies based on RET show strong robustness in the face of noise.
With the increase in noise level, the fail rate of all the tested strategies increases gradually.
In most cases, the baselines show a higher fail rate than our strategies. The robustness under noisy perturbation of Bi-NEAT and GS-NEAT are basically the same, the difference between the two is less than 0.4\%.
In the task with the low noise level, our strategies have a fail rate of one, as compared to dozens for the baselines.
However, in a few cases with high noise levels, all the strategies are unable to achieve results.

\section{Discussion}
In general, with the same fitness number of population, Bi-NEAT and GS-NEAT show better performance by ending up with fewer generations than NEAT for the symbolic logic, continuous control, and 2D gaming as the benchmark environments in this study.
Our proposed strategies are also superior in the tested tasks with incremental noisy observation.
We conclude than they are robust in the face of noise attacks, able to deal easily with sparse and noisy data.

More interestingly, the performance nuances of Bi-NEAT and GS-NEAT in different tasks also attracted our attention.
It is clear that Bi-NEAT is better than GS-NEAT in all tasks without noise.
Our preliminary conclusion is that evolutionary speed is affected by the $fitness$ landscape of different tasks, because the local peak of the landscape is usually small and sharp, as implied by the process data.
Another interesting point we observed is that GS-NEAT usually fares better than Bi-NEAT in the noise test. 
Further efforts could be performed to investigate the underneath mechanism and theoretical bounds. 

\section{Conclusion}

This paper introduced two specific evolutionary strategies based on RET for NE, namely Bi-NEAT and GS-NEAT.
The experiments with logic gates, Cartpole, and Lunar Lander show that Bi-NEAT and GS-NEAT have faster evolutionary speeds and greater stability than NEAT and FS-NEAT (baselines).
The noise test in Cartpole also shows stronger robustness than the baselines.

The influence of evolutionary speed, stability, and robustness of the whole strategy on the location selection of new topology nodes~\cite{yang2018learning} is worth further study on biological systems.
An assumption to validate is that this location selection can be adaptive vis-a-vis the landscape~\cite{ooi2018controllability} of generation.

\section*{Acknowledgments}
This work was initiated by Living Systems Laboratory at King Abdullah University of Science and Technology (KAUST) lead by Prof. Jesper Tegner and supported by funds from KAUST. Chao-Han Huck Yang was supported by the Visiting Student Research Program (VSRP) from KAUST.

\bibliographystyle{IEEEtran}
\bibliography{main}

\section*{Supplementary Materials}

\subsection{Open Source Library}
The codes and configurations are available in the  \href{https://github.com/HaolingZHANG/ReverseEncodingTree}{\textbf{Github}}.
This library has been improved and upgraded on \textbf{neat-python}~\cite{neatpython}.
By inheriting the \textbf{Class}.\textit{DefaultGenome}, the global individual, \textbf{Class}.\textit{GlobalGenome}, is realized.
The specific evolutionary strategies, like Bi-NEAT and GS-NEAT, inherit the \textbf{Class}.\textit{DefaultReproduction}, named \textit{bi} and \textit{gs} in the \textit{evolution/methods} folder.

In addition, we have created guidance models for our strategies, named \textit{evolutor}, in the \textit{benchmark} folder.
Our strategies can be used as independent algorithms for multi-modal search and as candidate plug-in units for other algorithms.

\subsection{Additional Results in the Noise Experiment}
In the noise experiment, the most important indicator is fail rate.
Some minor results, like the average and standard deviation of end generation, are also valuable.

\begin{figure}[htbp]
\centerline{\includegraphics[width=0.80\columnwidth]{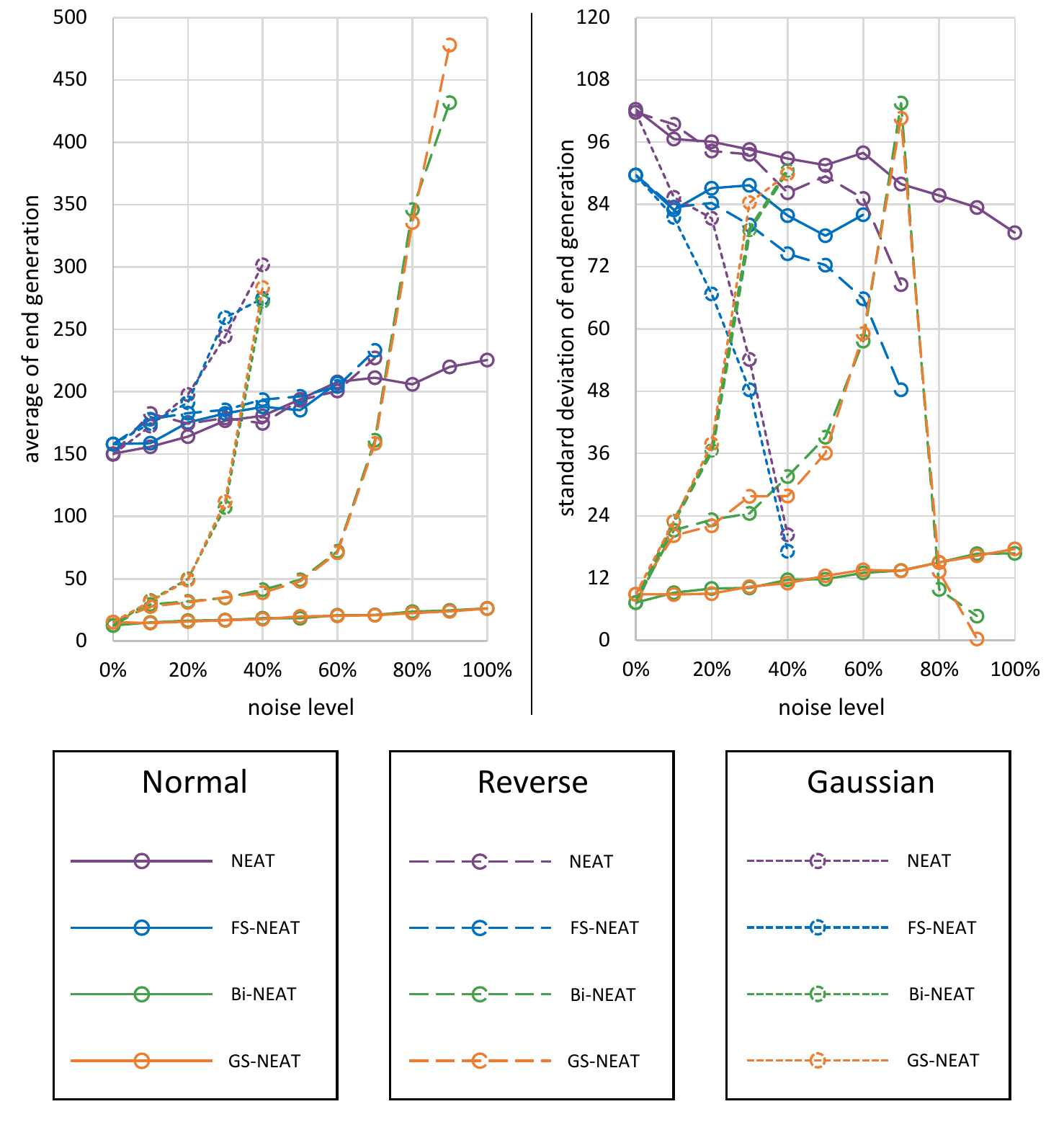}}
\caption{Avg.gen and StDev.gen in the noise experiments.}
\label{fig:additional}
\end{figure}

As shown in Fig.~\ref{fig:additional}, the average of end generation in each strategy increases with the increase in noise level.
Although the fail rates of our strategies are still low in the case of high noise levels, they need more generations to reach the fitness threshold.
The results from standard deviation describe the evolutionary difference between the baselines and our strategies.
Under noise attacks, the baselines will be unable to train, and will cause our strategies to delay achieving the requirements.

\subsection{Visualization of the Evolutionary Process}
RET is not only applicable to the field of NeuroEvolution, but can also be combined with other algorithms for tackling complex tasks.
Here we compare the evolution of RET and other well-accepted evolutionary strategies, to describe the evolutionary difference in the maximum or minimum position finding under the landscape.

The function landscapes, such as Rastrigin~\cite{hoffmeister1990genetic}, have potential patterns.
These potential patterns will determine the effect of the algorithm to some extent.

\begin{figure}[htbp]
\centerline{\includegraphics[width=1\columnwidth]{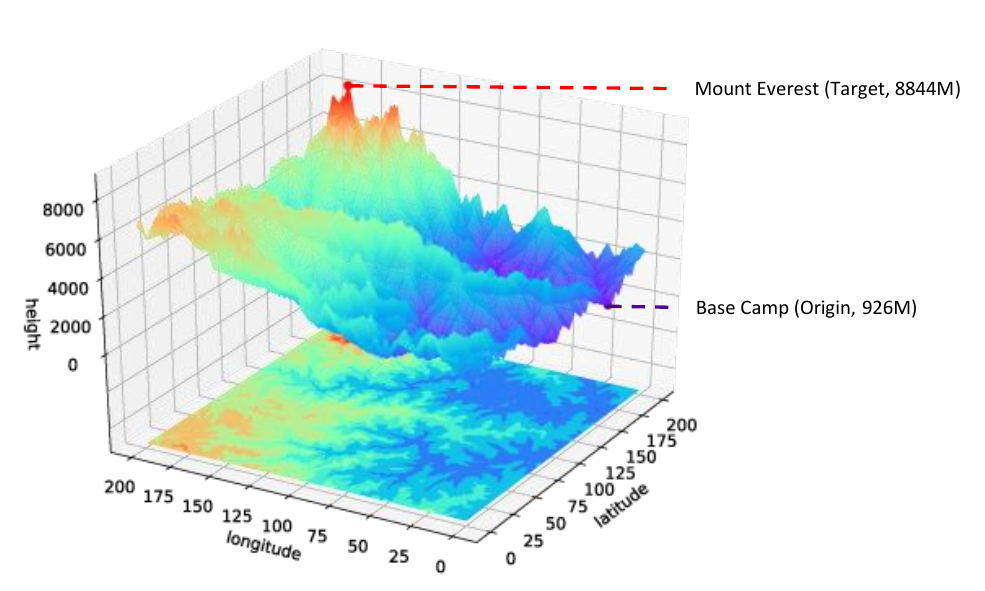}}
\caption{Landscape of Mount Everest with $200 \times 200$ points in $4 \times 4$ DEM.}
\label{fig:mount_everest}
\end{figure}

However, the landscape of the task built by NE is discrete.
After completing the experiment to find the minimum value of the Rastrigin function, we use the visualized 3D model \cite{gruen2002high} of Mount Everest.
The data set is from \textbf{Geospatial Data Cloud}~\footnote{\url{http://www.gscloud.cn/}}, $4 \times 4$ DEM around the Mount Everest, with $3600 \times 3600$ points.
Here, we compress $3600 \times 3600$ points into $200 \times 200$ points as the final discrete data, see Fig.~\ref{fig:mount_everest}.

\begin{figure*}[t]
\centerline{\includegraphics[width=2\columnwidth]{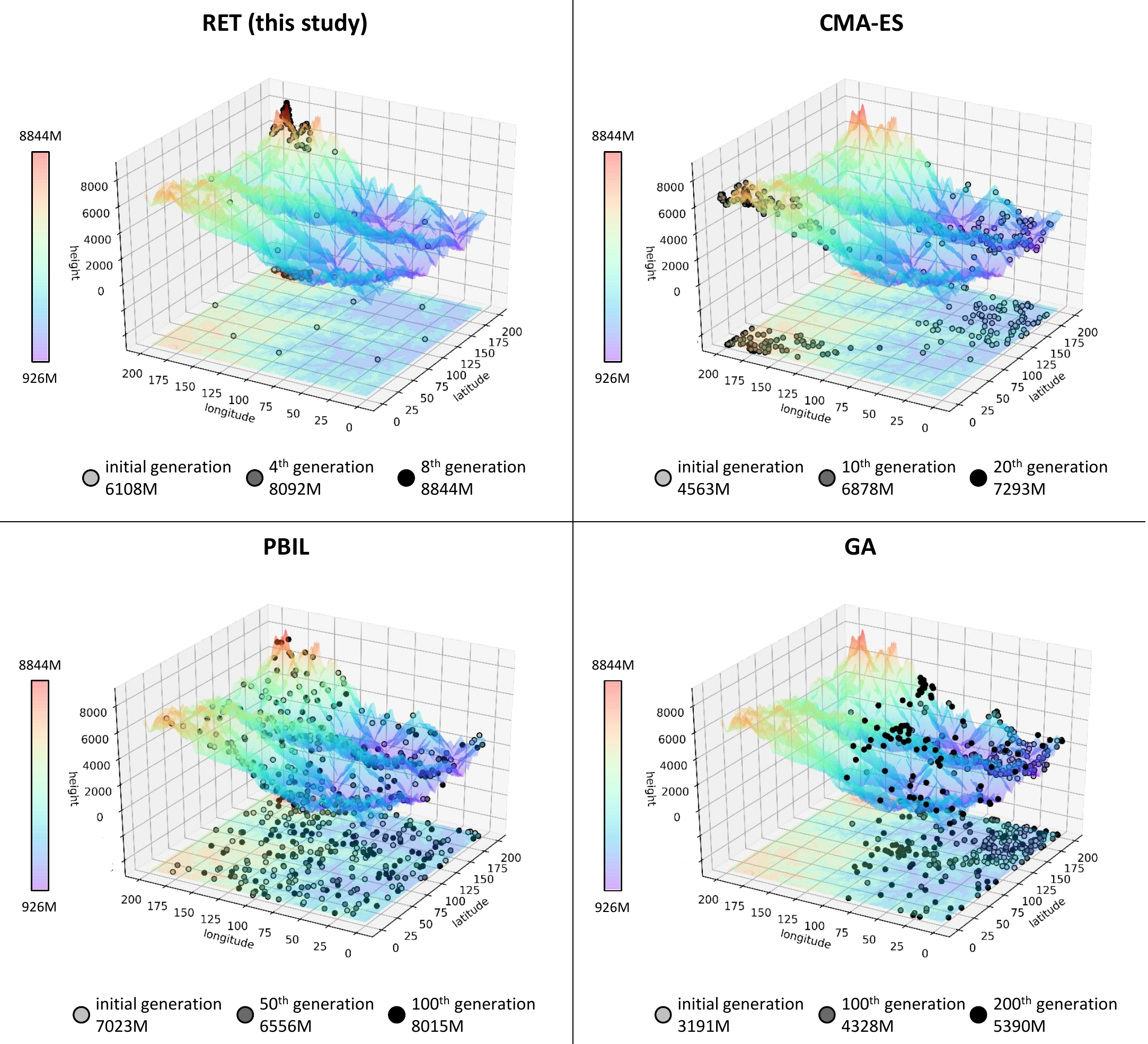}}
\caption{Using evolutionary strategies to find Mount Everest (8844m) in the landscape. 
}
\label{fig:evolution_vision}
\end{figure*}

The evolutionary process finding Mount Everest by different evolutionary strategies is shown in Fig.~\ref{fig:evolution_vision}.
The Mount Everest landscape with CSV format, named \textbf{mount\_everest.csv}, in \textit{benchmark/dataset} folder of our library.

\end{document}